\documentclass[11pt]{article}

\usepackage{arabtex}
\usepackage[utf8]{inputenc}
\usepackage{utf8}
\setcode{utf8}

\usepackage[final]{acl}

\usepackage{times}
\usepackage{latexsym}

\usepackage[T1]{fontenc}
\usepackage{microtype}

\usepackage{inconsolata}

\usepackage{graphicx}
\usepackage{xcolor}
\usepackage{soul}

\usepackage{hyperref}
\usepackage{caption}
\usepackage{subcaption}
\usepackage{enumitem}
\usepackage{paralist}
\usepackage{tabularx}
\usepackage{tcolorbox}
\usepackage{cleveref}

\newcommand{\sys}{\emph{Qayyem}}

\title{Qayyem: A Real-time Platform for Scoring Proficiency of Arabic Essays}

\author{
Hoor Elbahnasawi, Marwan Sayed, Sohaila Eltanbouly, Fatima Brahamia, Tamer Elsayed \\
Computer Science and Engineering Department, Qatar University \\
\texttt{\{he2006903, me2104862, se1403101, fb2205644, telsayed\}@qu.edu.qa}
}

\begin{document}
\maketitle
\begin{abstract}
Over the past years, Automated Essay Scoring (AES) systems have gained increasing attention as scalable and consistent solutions for assessing the proficiency of student writing. Despite recent progress, support for Arabic AES remains limited due to linguistic complexity and the scarcity of large publicly-available annotated datasets. In this work, we present \sys, a Web-based platform designed to support Arabic AES by providing an integrated workflow for assignment creation, batch essay upload, 
scoring configuration, and per-trait essay evaluation. \sys{} abstracts the technical complexity of interacting with scoring server APIs, allowing instructors to access advanced scoring services through a user-friendly interface. The platform deploys a number of state-of-the-art Arabic essay scoring models with different effectiveness and efficiency figures.
%, and we provide a detailed analysis of their effectiveness and inference efficiency.
\end{abstract}

\definecolor{softgreen}{RGB}{210,245,210}
\definecolor{softyellow}{RGB}{255,245,180}
\definecolor{softred}{RGB}{245,210,210}

% \sethlcolor{softgreen}
% \hl{Hoor}

% \sethlcolor{softyellow}
% \hl{Marwan}

% \sethlcolor{softred}
% \hl{Sohaila}

\section{Introduction } 
% \sethlcolor{softyellow} \hl{Marwan}

Automated Essay Scoring (AES) systems provide a scalable solution that supports teachers in academic writing assessments. 
%evaluations. 
They typically adopt holistic scoring, which assigns a single overall quality score to the essay \cite{npcr, pairwise-holistic-2025},
% representing the overall quality of the essay
or trait-specific scoring, which evaluates it on distinct writing traits, such as grammar or organization \cite{manyhands,ormerod2022mapping}. 
%Building on this, 
Developing such systems typically follows one of two paradigms: prompt-specific and cross-prompt. 
% The dominant approach, 
\textit{Prompt-specific} AES involves training and testing models on essays written 
%in response to 
for the same prompt, often achieving high performance due to the model’s specialization \cite{taghipour-ng-2016-neural,dong-etal-2017-attention}. 
In contrast, \textit{cross-prompt} AES models are tested on unseen prompts,
% that are unseen during training, 
%In contrast, \textit{cross-prompt} AES seeks to develop models that generalize across different writing prompts, 
enabling realistic and broader applicability, but presenting greater challenges due to increased topical variability \cite{ridley2021automated}. 

Despite the growing effort toward developing generalizable AES systems for English essays, research on Arabic AES remains limited, mainly due to 
%This gap can be partially attributed to 
the scarcity of large-scale Arabic datasets, which has constrained both methodological progress and system development. As a result, Arabic AES tools remain underexplored, particularly in terms of deployable software that supports realistic academic assessment workflows. 
Existing Arabic writing technologies primarily focus on writing assistance, such as Qalam,\footnote{\url{https://qalam.ai/en}} rather than automated essay scoring. 
%Within Arabic AES, 
ARWI \cite{chirkunov-etal-2025-arwi} represents the only known deployed software that targets essay scoring; however, it operates in a prompt-specific setting and does not support multi-trait scoring. 
% These limitations restrict its applicability in educational contexts with diverse writing topics and assessments that require structured, rubric-based scoring.
% where assignments vary across topics and assessments require structured, rubric-based scoring.

In this work, we introduce \sys,\footnote{Pronounced in Arabic as {\small ``\<قَيّم> ''} which means "grade" as a verb and "valuable" as a noun in English.} \emph{the first} cross-prompt multi-trait Arabic AES platform designed for deployment in academic settings. \sys{} is delivered through a web-based interface and supports a comprehensive assessment workflow. It allows teachers to create writing assignments, define custom rubrics, upload student essays, select specific writing traits for evaluating them, then 
%It then enables them to 
choose from a set of scoring models on a per-trait and/or per-assignment basis.

\sys{} supports both fully automated scoring and decision-support scenarios, allowing human oversight when required. It is designed to generalize to new assignments without retraining, addressing a key limitation of prompt-specific AES systems. The models provided within \sys{} are trained on LAILA, the largest available multi-trait Arabic AES corpus \cite{bashendy2025lailalargetraitbaseddataset}, enabling robust cross-prompt 
performance across diverse academic writing prompts. 

\begin{figure*}[t]
    \centering
    \includegraphics[width=\textwidth]{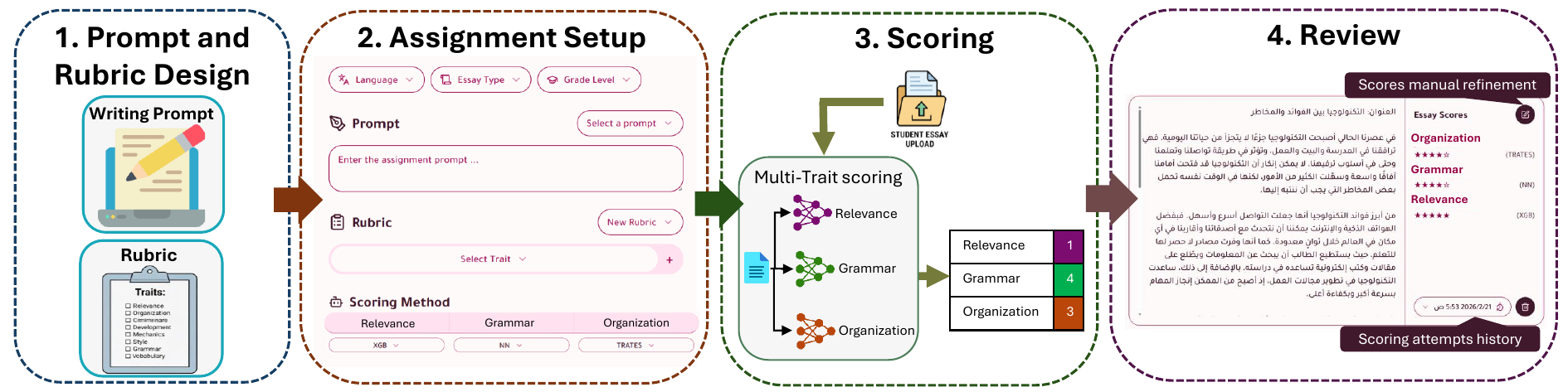}
    \caption{\sys{} workflow: 
    %illustrating 4 stages: 
    (1) Prompt and Rubric Design, where the writing prompt and scoring rubric are defined; (2) Assignment Setup, where scoring traits and evaluation settings are configured; (3) Scoring, where essays are auto-evaluated across selected traits; and (4) Review, where results are inspected and optionally refined.}
    \label{fig:wfd}
\end{figure*}

Our contribution is four-fold: 
% In summary, this work makes the following contributions: 
\begin{inparaenum}[(1)] 
\item we introduce \sys{},\footnote{\url{https://qayyem.qu.edu.qa/}}
% \footnote{A demo video is available at \url{https://youtu.be/LQwd-fLg8PI}.} 
the first cross-prompt multi-trait Arabic AES online system, 
%designed for realistic deployment in educational settings, 
\item we provide a Web-based interface supporting end-to-end assessment workflows for realistic educational settings,  
%, including assignment creation, rubric definition, model selection, automated scoring, and human-in-the-loop review, 
\item we deploy a range of 
%feature-based and 
state-of-the-art (SOTA) Arabic AES models, and
\item we provide public API access to the deployed models, enabling developers and researchers to query 
them 
%and demo our system 
directly.
%, and 
%\item we demonstrate the system’s effectiveness using LAILA, the largest available multi-trait Arabic AES corpus, showing robust cross-prompt performance across diverse writing prompts.
\end{inparaenum}

\section{Related Work} 
% \sethlcolor{softyellow}
% \hl{Marwan}

Automated writing support systems assist learners by providing feedback or evaluative scores.
% In this section, we review Arabic AES efforts and automated grading systems for other languages, finishing with work on Arabic MSA.
This section reviews related work on Arabic AES and automated support tools for different languages. 
% , followed by a discussion of efforts focused on Modern Standard Arabic (MSA).

% \subsection{Writing Assistance Tools}
% Existing writing assistance tools, including platforms like Grammarly,\footnote{\url{https://www.grammarly.com}} Wordtune,\footnote{\url{https://www.wordtune.com}} and QuillBot,\footnote{\url{https://quillbot.com}} which provide automated suggestions for grammar and style. Research prototypes such as ArgRewrite \cite{zhang-etal-2016-argrewrite} offer targeted support by automatically extracting and classifying sentence-level changes between drafts to explain revision purposes. This system focuses exclusively on the revision process rather than providing automated quality scoring or summative grades. 

% Additionally, the Automated Writing Evaluation (AWE) system \cite{litman2022automated} facilitates self-monitored revision using NLP techniques to distinguish between surface-level and content-altering revisions. The system visualizes these changes through interactive revision maps and distribution charts, prioritizing process-oriented feedback over performance-based scoring, providing multi-aspect feedback for iterative refinement.

\subsection{Automated Essay Scoring for Arabic}

Early Arabic AES relied on rule-based methods and feature engineering \cite{alqahtani-alsaif-2020-automated,alsanie2022automatic}, which were limited by manual feature design and struggled to generalize across prompts.
% , typically using traditional models like support vector regression with lexical, syntactic, and semantic features. 
% These approaches were limited by feature design and struggled to generalize across prompts.
% Some work applied text similarity techniques to align student essays with reference answers \cite{abdeljaber2021automatic,9498119,al2019automated,alobed2021automated,azmi2019aaee}, which captured prompt relevance but relied on shallow signals and rarely addressed cross-prompt evaluation.
% Recent studies shifted to neural networks and pretrained transformers. 
Recently, AraBERT has become a popular backbone for Arabic AES, where fine-tuning achieved strong performance \cite{ghazawi2024automated}, enhancements with handcrafted features improved relevance assessment \cite{machhout2024enhanced}, and parameter-efficient tuning further optimized AraBERT-based AES \cite{mahmoud2024automatic}. In addition, \citet{sayed2025feature} introduced 816 handcrafted features spanning surface, lexical, syntactic, readability, and semantic aspects, and used them to study cross-prompt AES.
More recently, large language models (LLMs) like ChatGPT and LLaMA have been evaluated for Arabic under zero-shot, in-context, and fine-tuning settings \cite{ghazawi2025well}.
% highlighting both their potential and challenges. 
Alongside these advances, 
% the TAQEEM 2025 shared task dataset \cite{taqeem2025}, comprising 1,265 essays across four prompts, was later incorporated into the larger LAILA corpus \cite{bashendy2025lailalargetraitbaseddataset}, which expands the benchmark to 7,859 essays across eight prompts with holistic and trait-level annotations, establishing the first large-scale standardized benchmark for Arabic AES.
\citet{bashendy2025lailalargetraitbaseddataset} introduced LAILA, a large-scale dataset comprising 7,859 essays across eight prompts with holistic and trait-level annotations, establishing the first large-scale standardized benchmark for Arabic AES.

% the TAQEEM 2025 shared task dataset \cite{taqeem2025} released 1,265 essays across four prompts, providing a standardized benchmark to support more rigorous and reproducible evaluation of Arabic AES systems.

% in robustness, interpretability, and cross-prompt generalization.

\subsection{Writing Support Tools}

Several automated writing support systems combine scoring with feedback. Criterion\textsuperscript{SM}~\cite{burstein2003criterionsm} is a commercial, cross-prompt AES system that produces holistic scores and feedback on grammar, usage, style, and discourse.
% , but its models are proprietary. 
Among non-commercial systems, eRevise+RF~\cite{liu-etal-2025-erevise} is a prompt-specific writing evaluation system that provides formative feedback;
% guiding formative instruction by evaluating student revisions, 
however, its models and data are not publicly released. 
% EssayCBM~\cite{chaudhary2025essaycbm} uses a rubric-aligned concept bottleneck approach, predicting scores for multiple concepts such as thesis clarity, evidence use, and organization, 
% Grammar, Vocabulary, Sentence Variety, Critical Thinking, and Fluency, 
% then aggregating them into a final grade. This provides actionable concept-level feedback and supports human-in-the-loop adjustments. The system uses multiple encoder backbones and is deployed as an interactive web application.
EssayCBM~\cite{chaudhary2025essaycbm} uses a rubric-aligned concept bottleneck model to predict concept-level scores and aggregate them into a final grade, providing actionable feedback with human-in-the-loop support via an interactive web application.

For other languages, \citet{hirao2020automated} introduced a publicly available prompt-specific system for non-native Japanese learners, generating both holistic and trait-specific scores with a BERT-based model.
% ; the system and code are publicly available.
IFlyEA~\cite{gong2021iflyea} is a cross-prompt Chinese AES system that analyzes grammar, rhetoric, and discourse analysis, providing explainable scoring, feedback, and visualizations.

For Arabic, ARWI~\cite{chirkunov-etal-2025-arwi} is the first publicly available prompt-specific AES system, supporting grammar checks and CEFR-aligned scoring.
% for Modern Standard Arabic (MSA). 
% It is a prompt-specific system that includes grammar checks and CEFR-aligned scoring,
% and a curated prompt database. 
While foundational for research, it lacks multi-trait scoring and cross-prompt generalization, limiting applicability in diverse educational settings.
% contexts with diverse writing topics and assessments.
% and assessments that require structured, rubric-based scoring. 
% \sys{} addresses these limitations by providing a range of cross-prompt models that support multi-trait scoring, enabling generalization to unseen writing prompts.
\sys{} addresses these limitations by providing cross-prompt models that support multi-trait scoring and generalization to unseen prompts.

\begin{figure*}[t]
    \centering
    \includegraphics[width=\textwidth]{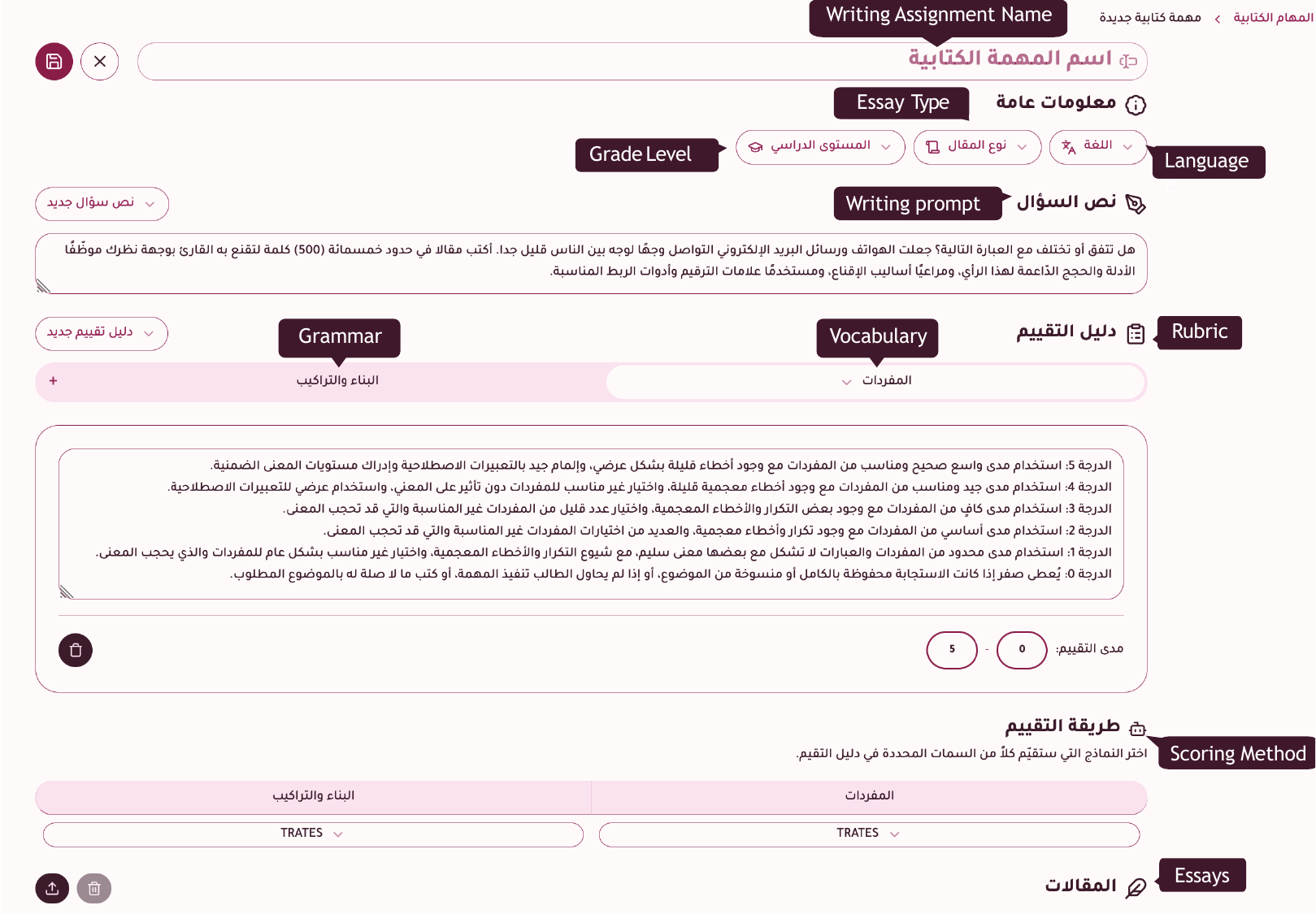}
    \caption{The assignment setup interface in Arabic, with English header translations. English translations of the prompt and rubric are provided in Appendix~\ref{rubric-examples}.}
    \label{fig:assignment-screenshots}
\end{figure*}

\begin{figure*}
    \centering
    \includegraphics[width=\textwidth]{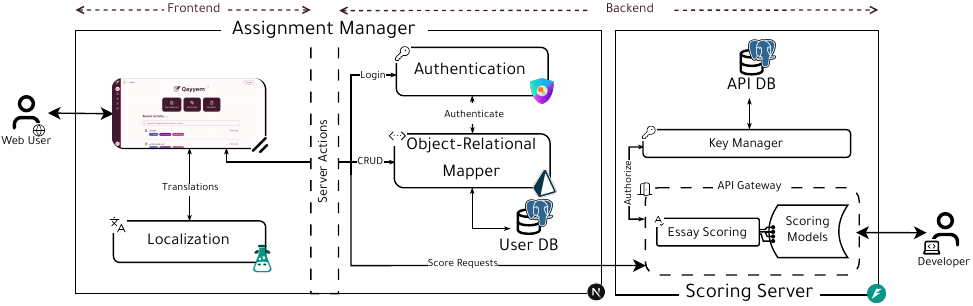}
    \caption{\sys's High-level Architecture Diagram.}
    \label{fig:system_design_diagram}
\end{figure*}

\section{System Workflow} 
% \sethlcolor{softgreen}

This section presents the overall workflow of \sys{}, as depicted in \autoref{fig:wfd}. 
%, and illustrate its application to Arabic AES. 
%The system workflow is depicted in \autoref{fig:wfd}. 
%To use \sys{}, 
The user first defines the writing prompts and rubrics (\S\ref{sec:stage1}), which are incorporated into the assignment setup along with other supplementary information (\S\ref{sec:assignment}). Essays are then uploaded and evaluated for the selected traits by the chosen scoring models (\S\ref{sec:scoring}). The resulting scores can be reviewed and, if needed, manually refined by the user (\S\ref{sec:review}).

% Essays are then uploaded and initially scored for the selected traits, providing preliminary results that can be reviewed and refined by the user.

% Each of these steps is described in detail in the following sections.

% The rest of the section explains how the system operates holistically, from the creation of prompts, rubrics, and assignments to the scoring and review process.

% \begin{figure*}[ht]
%     \centering
%     \includegraphics[width=\textwidth]{figures/Qayyem workflow diagram.v2.png}
%     \caption{\sys's Workflow Diagram}
%     \label{fig:wfd}
% \end{figure*}

\subsection{Prompt and Rubric Design} \label{sec:stage1}
The first step in designing a writing assignment %scoring any set of essays 
is to create the \emph{writing prompt} and the corresponding \emph{trait rubrics}. The writing prompt refers to the instructions and expected topic to be discussed in the essays. 
% The rubric is a list of grading criteria for a specific trait used to evaluate the essays. Each trait is accompanied by a score range and a description of the criteria used to assign the scores within the mentioned range.
The rubric defines the scoring criteria for evaluating a specific writing trait over a corresponding score range. 
%including the corresponding score range and the description of the criteria for assigning scores within that range.
Adding prompts and rubrics on \sys{} enables their reuse in subsequent assignments. An instructor may, for example, have a common rubric to score 8th-grade student essays, but the prompt changes for different assignments. 
% To create a prompt or rubric on \sys{}, a user may use the "New Prompt"/"New Rubric" quick action button available on the home page or the "Add New" button on the prompts' page or rubrics' list page. 
% Examples of writing prompts and scoring rubrics are provided in Appendix~\ref{rubric-examples}.

\subsection{Assignment Setup} \label{sec:assignment}
% Next, the user builds the assignment by including all of its elements, including: \textbf{General Information}, the language and type of the essays (e.g, persuasive, explanatory, or argumentative), the grade level at which the essays will be evaluated, and a title for the assignment. Next, a \textbf{Writing Prompt}, a \textbf{Rubric} (as defined in the subsection above), and the \textbf{Scoring Methods}, a mapping of which scoring models to be used for evaluating the selected traits. 
% % To create an assignment on \sys, a user may use the "New Assignment" quick action button available on the home page or the "Add New" button in the assignments' list page.

% After specifying the title and general assignment information, they may use any of their saved prompts or rubrics created in the previous step, edit, or create new ones on the spot. Then, the user will be able to choose which models will score which of the traits they had specified in the rubric. Each model has a 0-5 star rating of performance and a short description to give the user a general idea of what each model is. For the convenience of the user, each trait has its own assigned default model to score it, hand-picked by the developers based on its performance on our closed-source dataset for that specific trait.

Figure~\ref{fig:assignment-screenshots} illustrates the assignment creation interface (in Arabic, with English header translation). 
% The user constructs an assignment 
The assignment is constructed by specifying several needed information.
%all of its components. 
%through a unified interface. 
This first includes defining the \textbf{general information}: title, language of the assignment, essay type (e.g., persuasive, explanatory, or argumentative), and target grade level. 
Additionally, a \textbf{writing prompt} and \textbf{trait rubrics} 
%that specify the evaluation traits to be assessed 
%need to be added to the assignment. Prompts and rubrics can be 
are either selected from previously saved ones, edited, or newly created, allowing for flexible reuse and customization. Finally, 
\textbf{scoring models} are selected, one per designated trait.
%the \textbf{scoring methods} are specified, defining the mapping between selected traits and the scoring models used for their evaluation.
% Finally, the user specifies the \textbf{Scoring Methods}, which define the mapping between selected traits and the scoring models used for their evaluation. 
Each scoring model is accompanied by a short description and a 0-5 star performance rating to provide users with an overview of its characteristics. For usability, each trait is automatically associated with a default model, selected based on its performance.
%on a closed-source validation set.

\subsection{Scoring} \label{sec:scoring}
% After creating an assignment, 
After an assignment is created, essays can be uploaded in batches using a predefined format, ensuring consistent and efficient processing. The user can then select essays to be assessed and traits to be scored. 
%based on the defined rubrics.
% specifies the rubric-defined traits to score. 
The system processes the selected essays and returns scores in real time, as illustrated in \autoref{fig:scoring-view} in \autoref{appendix:scoring-interface}, with all results stored to allow optional review or refinement at any time. 
% Users may also perform multiple scoring runs using different models or assignment configurations, with each run tracked independently for optional review or refinement.
Multiple scoring runs can be performed using different models, 
%or assignment configurations, 
with each run tracked independently.
% for optional review or refinement.

\subsection{Review} \label{sec:review}
The Review interface allows inspection and management of previously generated scoring results. For each essay, it displays the full essay text alongside evaluated traits, scoring models, and assigned scores. \sys{} maintains a history of all scoring attempts, which can be browsed, compared, or deleted as needed, enabling traceability across different models and assignment configurations. Trait scores can also be \emph{manually refined} when necessary.
% Users may also manually refine trait scores. 
%The interface supports side-by-side inspection of multiple scoring runs to facilitate comparison. 
Finally, a summary report can be generated with the finalized scores for convenient reference.% or export.
% Finally,  users can generate a report summarizing the finalized scores for convenient reference or export.

\section{System Design and Implementation} 
% \sethlcolor{softgreen}
% \hl{Hoor}

% This section presents the architecture of \sys{}, illustrated in \autoref{fig:system_design_diagram}, which consists of two main components.
% %the assignment manager and the scoring server. 
% The \textbf{assignment manager} provides a Web interface for creating and storing assignments, managing user data, and coordinating communication with the scoring server. The \textbf{scoring server} hosts the essay scoring models and returns real-time scoring progress through its API.

This section presents the architecture of \sys{}, illustrated in \autoref{fig:system_design_diagram}, which consists of two main components: the assignment manager (\S\ref{subsec:assignment_manager}) and the scoring server (\S\ref{subsec:scoring_server}). The assignment manager provides the Web interface for managing assignments, while the scoring server hosts the models and returns the scores. The section then details the system's deployment (\S\ref{subsec:deployment}) and the specific scoring models used (\S\ref{subsec:scoring_models}).

\subsection{Assignment Manager} \label{subsec:assignment_manager}

The assignment manager is \sys{}'s primary interface, presented as a website through which instructors interact with the system. As shown in \autoref{fig:system_design_diagram}, it is responsible for managing assignments and coordinating communication between users and the scoring server. 
%The assignment manager 
It supports both Arabic and English user interfaces (UIs) and allows users to create, edit, and delete prompts, rubrics, and assignments, providing a centralized workspace for essay grading.

The assignment manager is implemented as a full-stack Web application using Next.js,\footnote{\url{https://nextjs.org/}} enabling tight integration between frontend interactions and backend logic. Actions performed through the Web interface are handled via server-side actions that enforce access control and trigger appropriate system workflows. This design allows instructors to access the essay scoring functionality without interacting directly with the scoring server or its API.

On the backend, the assignment manager handles user authentication, authorization, and persistent data management. 
%Authentication is integrated 
%into the application 
%to ensure that only authorized users can access assignments and submit scoring requests. 
Access to the system requires a valid user account.
% \footnote{%As the platform is currently closed for open registration,
% Demo credentials are provided for evaluation: {Email: \texttt{demo@acl.com} Password: \texttt{demoAcl2026!}}.} 
Assignment-related data, including users, prompts, rubrics, assignments, and scores, are stored in a PostgreSQL database\footnote{\url{https://www.postgresql.org/}} and accessed via an object-relational mapping layer implemented with Prisma,\footnote{\url{https://www.prisma.io/}} ensuring a structured and consistent data model.

When essays are uploaded and submitted for scoring, the assignment manager sends an authorized scoring request to the scoring server through its API. 
% During scoring, the assignment manager tracks progress updates and reflects them in the user interface. 
During scoring, it tracks progress and updates the UI accordingly. Once 
%scoring is 
completed, scores are stored in the database and made available 
%to the user 
through the UI. %through the assignment manager.

\subsection{Scoring Server} \label{subsec:scoring_server}

% The scoring server is responsible for the main functionality of \sys{}, managing automatic essay scoring requests from website and developer users. This component was wrapped using FastAPI,\footnote{https://fastapi.tiangolo.com/} a simple framework for defining and managing the API endpoints. As the server initially starts up, it checks the centralized JSON configuration files that define which models are added to the server, which are active, which load on startup, and all other relevant information about the models (names, descriptions, performance ratings, etc). This allows for a very decoupled setup, where system admins have a single point of entry to adding or disabling models by specifying them in the files, since they all follow a unified interface class and interact with the API service via the same functions. 

% Additionally, the server tracks ongoing requests and sends real-time updates, using Server-Sent Events (SSEs), of scores as scoring models process them. Finally, an API key authorization system is set up, requiring an access request and rate limitations for any developers wishing to utilize the scoring models outside of the \sys{} website. The API keys for \sys{} are available for request via form.

% there needs to be a mention of the scoring interface and how we can add models and the deployment proccess needs to be explained also maybe a subsection?

The scoring server is responsible for the core functionality of \sys{}, handling automated scoring requests submitted by both website users and external developers. As shown in \autoref{fig:system_design_diagram}, it operates as a standalone backend service that exposes a unified API for accessing the available scoring models.

The scoring server is implemented using FastAPI\footnote{\url{https://fastapi.tiangolo.com/}} to 
%define and 
manage its API endpoints. Upon startup, it reads from centralized configuration files that specify (among others) the available scoring models, including which models to enable and load, 
%are enabled and loaded, 
%at initialization, 
along with their associated metadata. This %configuration-based 
approach allows system administrators to add or disable models without modifying the core server logic, as all models adhere to a shared interface.

During execution, the scoring server manages active scoring requests and provides real-time progress updates as essays are evaluated. These updates are streamed to clients using Server-Sent Events (SSEs). Once scoring is completed, the final results are returned through the API.

Access to the scoring server is protected through API key–based authorization. External requests are routed through an API gateway and require API keys, which are managed by a dedicated key manager, 
%as shown in \autoref{fig:system_design_diagram}, 
responsible for access control and rate limiting. The API supports both REST and Server-Sent Events (SSE) endpoints, enabling synchronous configuration queries and real-time streaming of scoring progress. 
%respectively. 
API keys are requested via a dedicated form.\footnote{\url{https://qayyem.qu.edu.qa/api-request}} Detailed API documentation, including endpoint specifications, request formats, and code examples
% in multiple languages, 
is available online.\footnote{\url{https://qayyem.qu.edu.qa/documentation}}

% API keys are required for external requests and are subject to access control and rate limitations. 
% External requests require API keys and are subject to access control and rate limitations. 
% % Developers may request API keys through a dedicated form.\footnote{\url{https://qayyem.qu.edu.qa/api-request}}
% API keys can be requested through a dedicated form.\footnote{\url{https://qayyem.qu.edu.qa/api-request}}
% Detailed API documentation, including endpoint specifications and usage guidelines, is available online.\footnote{\url{https://qayyem.qu.edu.qa/documentation}}

% there needs to be a mention of the scoring interface and how we can add models and the deployment proccess needs to be explained also maybe a subsection?
\subsection{Deployment} \label{subsec:deployment}

\sys{} is 
%currently 
deployed on a single Linux-based server 
%occupied 
with NVIDIA A10 GPU and AMD EPYC 74F3 24-core processor.
% \hl{[More specs here]}. 
NGINX\footnote{\url{https://nginx.org/}} is used as a reverse proxy to handle incoming requests and route traffic to the appropriate system components. The assignment manager 
%(website) 
is deployed as a persistent Node.js service using PM2,\footnote{\url{https://pm2.keymetrics.io/}} while the scoring server 
%(API) 
runs as a standalone backend service using Uvicorn.\footnote{\url{https://uvicorn.dev/}}

The scoring server relies on external configuration files to define the supported languages, grade levels, essay types, traits, and available scoring models. 
This 
%These configuration files allow 
allows the system to be adapted or extended without changes to the deployment setup, and enables controlled updates and maintenance. 
%Configuration metadata also specifies which scoring models 
%and traits are available to users and whether they 
%are active at runtime, enabling controlled updates and maintenance.
During deployment or restart, the scoring server loads the active configuration into memory, making the supported traits and models immediately available to the API and assignment manager. Models may be initialized at startup or activated on demand, depending on their configuration. This design allows the system to manage computational resources while remaining responsive to scoring requests.

As new scoring models emerge, they can be deployed by adding their corresponding configuration and model support files, without disrupting existing deployments. Previously-deployed models are retained 
%in the configuration and can be 
or disabled when no longer needed, ensuring consistency between the scoring server and the assignment manager, particularly with respect to database records and the tracking of historical scores, where stable model identifiers are required to preserve and reference past results.

\begin{table*}[]
\centering
\small
\begin{tabular}{llllllllll}
\hline
\textbf{Model} & \textbf{REL} & \textbf{ORG} & \textbf{VOC} & \textbf{STY} & \textbf{DEV} & \textbf{MEC} & \textbf{GRM} & \textbf{HOL} & \textbf{AVG} \\
\hline
RF & 0.331 & 0.609 & \ul{0.644} & 0.637 & 0.573 & 0.559 & 0.609 & \ul{0.682} & 0.581 \\
NN & 0.353 & 0.609 & 0.621 & 0.631 & 0.566 & 0.565 & 0.597 & 0.651 & 0.574 \\
XGB & 0.360 & \ul{0.645} & 0.641 & 0.641 & 0.583 & 0.577 & 0.619 & 0.679 & 0.593 \\
MOOSE & \ul{0.411} & 0.627 & 0.642 & \ul{0.649} & \ul{0.585} & \ul{0.586} & \ul{0.623} & 0.649 & \ul{0.597} \\
TRATES & \textbf{0.557} & \textbf{0.696} & \textbf{0.657} & \textbf{0.664} & \textbf{0.652} & \textbf{0.608} & \textbf{0.643} & \textbf{0.744} & \textbf{0.653} \\

\hline
\end{tabular}
\caption{
Average QWK performance over all prompts for each trait. \textbf{Bold} and \ul{underlined} values indicate the best and second-best performance per trait. 
% The last column shows the inference time in \textbf{seconds} for one essay.
}
\label{tab:result}
\end{table*}

\subsection{Scoring Models} \label{subsec:scoring_models}

The scoring models deployed on \sys{} are chosen to balance SOTA performance with inference efficiency, providing a diverse set of models for Arabic AES. 
The implementation details for all the models are provided in Appendix \ref{appendix:immplementation-detail}. We note that although the currently deployed models support Arabic, the system’s bilingual UI enables seamless integration of English AES models.

\paragraph{Feature-based Models}
Neural Networks (NN), Random Forest (RF), and XGBoost (XGB) were chosen as feature-based baselines for their strong performance among feature-based approaches \cite{li-ng-2024-conundrums,bashendy2025lailalargetraitbaseddataset,sayed2025feature},\footnote{XGB achieves second-best performance on LAILA.} and computational efficiency.
% These models leverage handcrafted features to capture various linguistic aspects of essays, 
%These models 
They leverage 816 handcrafted features introduced by \citet{sayed2025feature} to capture various linguistic aspects of essays,
providing robust and interpretable scoring while serving as efficient complements to the more computationally-intensive deployed models. %deployed on \sys. 
% \subsection{Feature Set}

% We adopt the 816 handcrafted features introduced by \citet{sayed2025feature} for Arabic AES, covering surface, readability, lexical, semantic, and syntactic aspects. To reduce noise, we performed feature selection based on Pearson and Spearman correlations \cite{li-ng-2024-conundrums}, retaining features whose absolute correlation with any trait exceeded a predefined threshold. For training the deployment models, we removed 22 features that require pre-trained transformer models for extraction. These features were found to contribute minimally to overall performance while increasing the scoring time; therefore, they were removed to improve efficiency.

\paragraph{State-of-the-Art (SOTA) Models}
We 
%reimplemented )
adapted 
two SOTA English AES models for Arabic. The first is TRATES~\cite{eltanbouly-etal-2025-trates}, which generates trait-specific features using an LLM (Fanar \cite{team2025fanar} in our adaptation), combined with linguistic features for scoring using a simple NN. 
%In our adaptation, we use Fanar \cite{team2025fanar}.
% , which is an open-source Arabic-centric LLM.
% with extraction optimized by querying all rubric-based questions for an essay in a single call. 
% Engineered features include length, readability, POS, complexity, sentiment, and prompt-specific features derived from the Arabic feature set.
The second is MOOSE~\cite{chen-etal-2025-mixture-ordered}, which combines 
%three experts 
mixture-of-experts
for trait prediction, pairwise essay ranking, and essay–prompt relevance (using AraBERT encoder \cite{antoun2020arabert} in our adaptation).
%via a mixture-of-experts. 
%For Arabic, we adapted MOOSE using AraBERT \cite{antoun2020arabert}.
% and the same set of Arabic handcrafted features \cite{sayed2025feature}.
%Both adaptations preserve the original methodology while ensuring suitability for 
%deployment on \sys{} for 
%Arabic.

% \begin{table*}[]
% \centering
% \begin{tabular}{llllllllll|r}
% \hline
% \textbf{Model} & \textbf{REL} & \textbf{ORG} & \textbf{VOC} & \textbf{STY} & \textbf{DEV} & \textbf{MEC} & \textbf{GRM} & \textbf{HOL} & \textbf{AVG} & \textbf{Time} \\
% \hline
% RF & 0.331 & 0.609 & \ul{0.644} & 0.637 & 0.573 & 0.559 & 0.609 & \ul{0.682} & 0.581 & 0.37\\
% NN & 0.353 & 0.609 & 0.621 & 0.631 & 0.566 & 0.565 & 0.597 & 0.651 & 0.574 & 0.26 \\
% XGB & 0.360 & \ul{0.645} & 0.641 & 0.641 & 0.583 & 0.577 & 0.619 & 0.679 & 0.593 & 0.50\\
% MOOSE & \ul{0.411} & 0.627 & 0.642 & \ul{0.649} & \ul{0.585} & \ul{0.586} & \ul{0.623} & 0.649 & \ul{0.597} & 1.05\\
% TRATES & \textbf{0.557} & \textbf{0.696} & \textbf{0.657} & \textbf{0.664} & \textbf{0.652} & \textbf{0.608} & \textbf{0.643} & \textbf{0.744} & \textbf{0.653} & 24.72\\

% \hline
% \end{tabular}
% \caption{
% Average QWK performance over the 8 prompts for each trait. \textbf{Bold} and \ul{underlined} values indicate the best and second-best performance per trait. The last column show the inference time in \textbf{seconds} for one essay.}
% \label{tab:result}
% \end{table*}

\section{Experimental Evaluation}
% \sethlcolor{softyellow}
% \hl{Marwan}

% In this section, we discuss the experimental setup for all of the models deployed on \sys, along with the dataset used, the implementation details, and the training setups.
In this section, we report and discuss 
the performance of the deployed scoring models.
%We first present the training dataset, 
%the dataset used to train all of the models, along with their performance.

%\subsection{Experimental Setup}
\paragraph{Experimental Setup}
%\subsection{Dataset and Evaluation}
% The limited availability of large-scale, standardized Arabic essay corpora has long constrained progress in Arabic AES. 
% In this study, 
To develop the deployed models, we utilize the newly-released dataset, LAILA~\cite{bashendy2025lailalargetraitbaseddataset}, 
% which constitutes a multi-trait annotated corpus 
designed to advance robust Arabic AES research. It contains 7,859 Arabic essays written by students in grades 10-12 across 8 prompts (5 persuasive and 3 explanatory).
%collected from 24 schools in Qatar under exam-like time-constrained conditions. 
The essays are annotated using a unified scoring rubric in 7 writing traits: Relevance (REL, 0–2), Organization (ORG, 0–5), Vocabulary (VOC, 0–5), Style (STY, 0–5), Development (DEV, 0–5), Mechanics (MEC, 0–5), and Grammar (GRM, 0–5), in addition to Holistic score (HOL) computed as the sum of all trait scores. 
% For model training, the released predefined cross-prompt splits are used.
% Annotation follows standardized rubrics from the Core Academic Skills Test (CAST) developed by the Qatar University Testing Center (QUTC)\footnote{\url{https://www.qu.edu.qa/sites/en_US/testing-center/TestDevelopment/cast}}, and employs a double-scoring procedure with adjudication to ensure annotation reliability. The scale, prompt diversity, and multi-trait nature of LAILA enable the analyses and modeling presented in this work.
% \subsection{Evaluation} \label{evaluation}
To evaluate the models, we use Quadratic Weighted Kappa (QWK)~\cite{cohen1968weighted}, a common measure for AES that assesses the agreement between the scores of two raters. All models are trained in a cross-prompt setting using the released splits of LAILA dataset. 

\begin{figure}
    \centering
    \includegraphics[width=\linewidth]{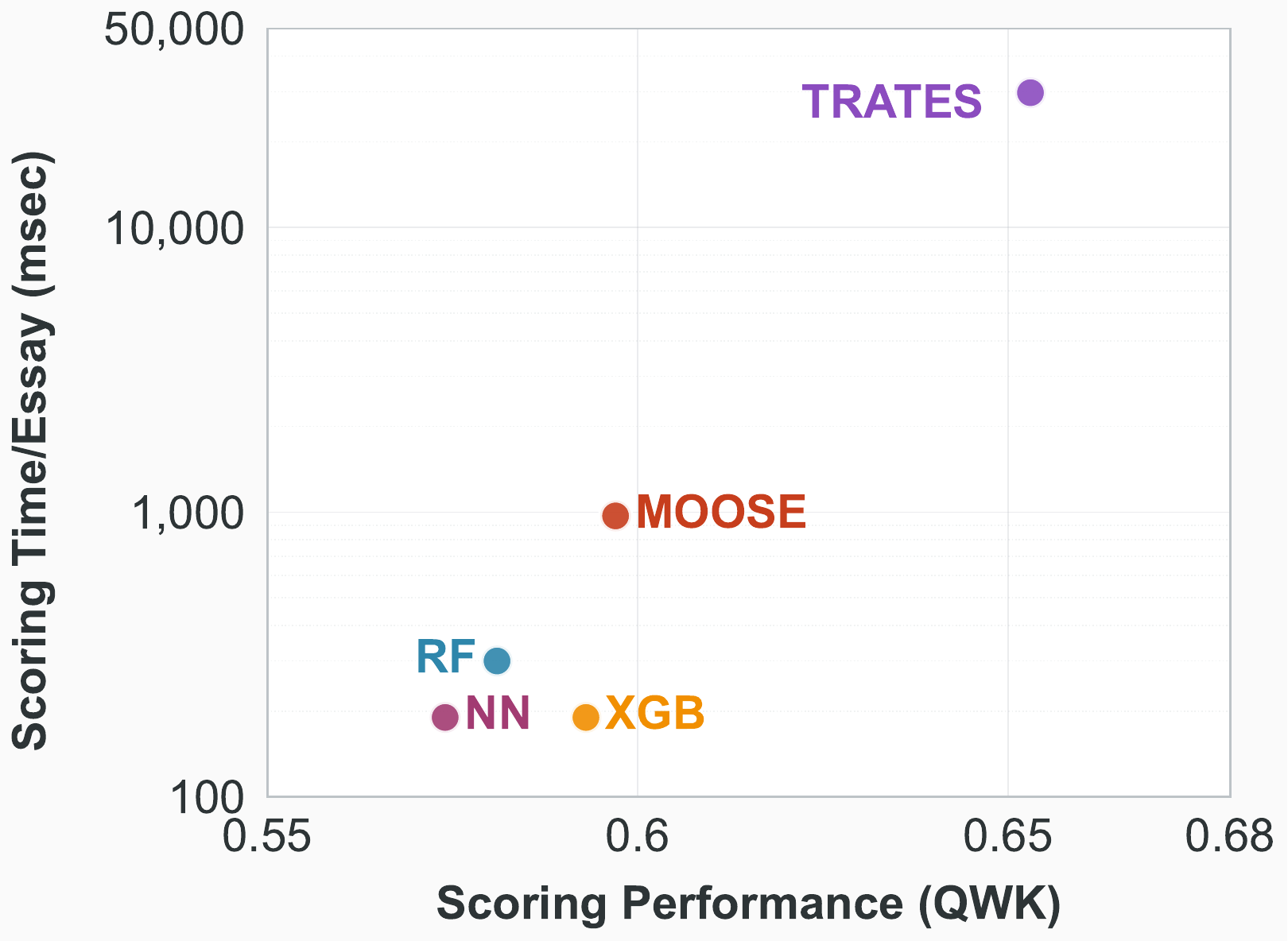}
    %\caption{Scoring effectiveness measured by QWK versus per-essay inference time in milliseconds on a logarithmic scale across the different scoring models.}
    \caption{Scoring effectiveness (in QWK) vs. per-essay inference time (in msec) across deployed models.}
    \label{fig:effectiveness_efficiency_from_zero}
\end{figure}

%\subsection{Results}
% \sethlcolor{softred}
% \hl{Sohaila}
% \usepackage[normalem]{ulem}

% \begin{table}[]
% \centering  
% \begin{tabular}{lrrr}
% \hline
% \textbf{\#Essays} & \textbf{10} & \textbf{50} & \textbf{100} \\
% \hline
% RF & 4.2 & 19.0 & 37.2 \\
% NN & 3.1 & 12.9 & 25.7 \\
% XGB & 5.6 & 25.7 & 50.0 \\
% TRATES & 243.4 & 1227.3 & 2472.0 \\
% MOOSE & 11.5 & 52.8 & 104.8 \\
% \hline
% \end{tabular}
% \caption{The inference time in \textbf{seconds} for each model for scoring 10, 50, and 100 essays}
% \label{tab:inference-time}
% \end{table}

\paragraph{Effectiveness} Table~\ref{tab:result} reports the average performance on LAILA over the 8 prompts for each writing trait. TRATES achieves the best overall performance across all traits, followed by MOOSE, highlighting the superiority of SOTA approaches over traditional feature-based models. Notably, TRATES outperforms MOOSE by 15 points on the relevance trait. 
%and exceeds the lowest-performing model by 23 points. 
For vocabulary and style, performance differences across models are relatively small, with a gap range of about 3.5 points. 
%between the highest- and lowest-ranked models. 
This gap increases to around 5 points for mechanics and grammar, and further widens to 9 points for holistic, development, and organization traits. However, these performance gains must be interpreted in light of the computational cost required to achieve them.

\paragraph{Effectiveness--Efficiency Trade-off}
Figure~\ref{fig:effectiveness_efficiency_from_zero} illustrates the scoring performance measured in QWK vs. average inference time per essay. Feature-based models are clustered in the low inference-time region, with NN, RF, and XGB requiring 0.2, 0.3, and 0.2 seconds per essay, respectively, exhibiting relatively lower QWK values than other models. 
%, and exhibiting average QWK values of 0.574, 0.581, and 0.593. 
MOOSE occupies an intermediate position, with an inference time of 1 second per essay and 
a bit higher QWK than feature-based models, 
%an average QWK of 0.597, 
reflecting its more complex architecture and increased computational requirements.
% compared to feature-based approaches. 
In contrast, TRATES lies at the extreme end of the inference-time spectrum, requiring 30 seconds per essay while achieving the best QWK.
%an average QWK of 0.653. 
This increased inference time is attributed to its trait-specific design, which involves a separate model call for each trait. In addition, it requires substantially greater computational resources, either through high-capacity GPUs for local inference or LLM-based API calls. 

% Overall, the figure highlights clear differences in computational cost and inference time across models with respect to their QWK scores.

This comparison guides users in selecting models for trait-level essay scoring, highlighting the trade-off between effectiveness and efficiency.

% Table \ref{tab:inference-time} reports the inference time required to score 10, 50, and 100 essays. FB models exhibit low inference times, ranging from 3 seconds for the NN to 5.6 seconds for XGB when scoring 10 essays. MOOSE requires more time due to its complex architecture, taking 11.5 seconds for 10 essays, and also requires greater computational resources (e.g., a GPU) for deployment. In contrast, TRATES is considerably slower, requiring approximately 4 minutes to score 10 essays. It is worth noting that TRATES is the only deployed trait-specific model, requiring a separate model call for each trait, whereas the other models adopt multi-task approaches that predict all traits in a single inference pass. In addition, TRATES requires substantially more computational resources for deployment, either through a more powerful GPU for local inference or via API calls to an LLM backend, which accounts for most of its inference time. 

% Although TRATES achieves superior performance, it suffers from lower efficiency due to its reliance on a large language model, resulting in significantly higher inference time than the other approaches.

% This comparison guides users in selecting models for trait-level essay scoring by highlighting the trade-off between effectiveness and efficiency.

% For the deployment models, we retrain all the models on the full dataset using the most common hyperparameters obtained from the individual folds. 

\section{Conclusion }
% \sethlcolor{softred}
% \hl{Sohaila}
In this paper, we introduced \sys{}, the first Arabic AES platform to support fully-automated essay scoring while providing instructors with a flexible and user-friendly interface. The platform allows instructors to create assignments, define writing prompts and trait-specific rubrics, and configure scoring methods. The current version of \sys{} supports five scoring models, enabling users to select and apply SOTA trait-based models according to their needs. 
% The system abstracts the technical complexity of interacting with scoring server APIs, allowing users without technical expertise to access advanced AES services seamlessly. 
By integrating real-time scoring, manual score refinement, and report generation, \sys{} provides a comprehensive solution for managing the full essay evaluation workflow. 

As future work, we plan to extend \sys{} to support additional Arabic scoring models,  incorporate English scoring models, integrate systems capable of generating formative feedback to students, support student user scenarios, and provide a tiered system for different types of users.

\section*{Acknowledgment}
The work of Sohaila Eltanbouly was supported by
GSRA grant\# GSRA12-L-0413-250111, and the work of the other authors was supported by NPRP grant\# NPRP14S-0402-210127, 
both from the Qatar National Research Fund (a member of Qatar
Foundation). The statements made herein are solely the responsibility of the authors.

% \sys{} is implemented with scalability in mind, allowing for easy integration of new scoring models by developers. 
% \section*{Acknowledgments}

% \newpage
\bibliography{custom}

\begin{thebibliography}{27}
\providecommand{\natexlab}[1]{#1}

\bibitem[{Alqahtani and Alsaif(2020)}]{alqahtani-alsaif-2020-automated}
Abeer Alqahtani and Amal Alsaif. 2020.
\newblock \href {https://aclanthology.org/2020.icon-main.24/} {Automated {A}rabic essay evaluation}.
\newblock In \emph{Proceedings of the 17th International Conference on Natural Language Processing (ICON)}, pages 181--190, Indian Institute of Technology Patna, Patna, India. NLP Association of India (NLPAI).

\bibitem[{Alsanie et~al.(2022)Alsanie, Alkanhal, Alhamadi, and Alqabbany}]{alsanie2022automatic}
Waleed Alsanie, Mohamed~I Alkanhal, Mohammed Alhamadi, and Abdulaziz~O Alqabbany. 2022.
\newblock Automatic scoring of {A}rabic essays over three linguistic levels.
\newblock \emph{Progress in Artificial Intelligence}, pages 1--13.

\bibitem[{Antoun et~al.(2020)Antoun, Baly, and Hajj}]{antoun2020arabert}
Wissam Antoun, Fady Baly, and Hazem Hajj. 2020.
\newblock \href {https://aclanthology.org/2020.osact-1.2} {{A}ra{BERT}: Transformer-based model for {A}rabic language understanding}.
\newblock In \emph{Proceedings of the 4th Workshop on Open-Source Arabic Corpora and Processing Tools, with a Shared Task on Offensive Language Detection}, pages 9--15, Marseille, France. European Language Resources Association.

\bibitem[{Bashendy et~al.(2026)Bashendy, Massoud, Eltanbouly, Albatarni, Sayed, Abir, Bouamor, and Elsayed}]{bashendy2025lailalargetraitbaseddataset}
May Bashendy, Walid Massoud, Sohaila Eltanbouly, Salam Albatarni, Marwan Sayed, Abrar Abir, Houda Bouamor, and Tamer Elsayed. 2026.
\newblock \href {https://aclanthology.org/2026.eacl-long.142/} {{LAILA}: A large trait-based dataset for {A}rabic automated essay scoring}.
\newblock In \emph{Proceedings of the 19th Conference of the European Chapter of the Association for Computational Linguistics (EACL 2026)}. Association for Computational Linguistics.

\bibitem[{Burstein et~al.(2003)Burstein, Chodorow, and Leacock}]{burstein2003criterionsm}
Jill Burstein, Martin Chodorow, and Claudia Leacock. 2003.
\newblock Criterionsm online essay evaluation: An application for automated evaluation of student essays.
\newblock In \emph{Proceedings of the Fifteenth Annual Conference on Innovative Applications of Artificial Intelligence (IAAI)}, pages 3--10. Association for the Advancement of Artificial Intelligence.
\newblock Reposted with permission on ETS.org.

\bibitem[{Chaudhary et~al.(2025)Chaudhary, Zhao, Zhang, Tse, Agrawal, Deng, and Liu}]{chaudhary2025essaycbm}
Kumar~Satvik Chaudhary, Chengshuai Zhao, Fan Zhang, Yung~Hin Tse, Garima Agrawal, Yuli Deng, and Huan Liu. 2025.
\newblock \href {https://arxiv.org/abs/2512.20817} {Essay{CBM}: Rubric-aligned concept bottleneck models for transparent essay grading}.
\newblock \emph{arXiv preprint arXiv:2512.20817}.

\bibitem[{Chen et~al.(2025)Chen, Tsai, Wei, Wang, Wang, and Huang}]{chen-etal-2025-mixture-ordered}
Po-Kai Chen, Bo-Wei Tsai, Shao~Kuan Wei, Chien-Yao Wang, Jia-Ching Wang, and Yi-Ting Huang. 2025.
\newblock \href {https://doi.org/10.18653/v1/2025.acl-long.884} {Mixture of ordered scoring experts for cross-prompt essay trait scoring}.
\newblock In \emph{Proceedings of the 63rd Annual Meeting of the Association for Computational Linguistics (Volume 1: Long Papers)}, pages 18071--18084, Vienna, Austria. Association for Computational Linguistics.

\bibitem[{Chirkunov et~al.(2025)Chirkunov, Alhafni, Qwaider, Habash, and Briscoe}]{chirkunov-etal-2025-arwi}
Kirill Chirkunov, Bashar Alhafni, Chatrine Qwaider, Nizar Habash, and Ted Briscoe. 2025.
\newblock \href {https://doi.org/10.18653/v1/2025.in2writing-1.2} {{ARWI}: {A}rabic write and improve}.
\newblock In \emph{Proceedings of the Fourth Workshop on Intelligent and Interactive Writing Assistants (In2Writing 2025)}, pages 11--18, Albuquerque, New Mexico, US. Association for Computational Linguistics.

\bibitem[{Cohen(1968)}]{cohen1968weighted}
Jacob Cohen. 1968.
\newblock Weighted kappa: nominal scale agreement provision for scaled disagreement or partial credit.
\newblock \emph{Psychological bulletin}, 70(4):213.

\bibitem[{Dong et~al.(2017)Dong, Zhang, and Yang}]{dong-etal-2017-attention}
Fei Dong, Yue Zhang, and Jie Yang. 2017.
\newblock \href {https://doi.org/10.18653/v1/K17-1017} {Attention-based recurrent convolutional neural network for automatic essay scoring}.
\newblock In \emph{Proceedings of the 21st Conference on Computational Natural Language Learning ({C}o{NLL} 2017)}, pages 153--162, Vancouver, Canada. Association for Computational Linguistics.

\bibitem[{Eltanbouly et~al.(2025)Eltanbouly, Albatarni, and Elsayed}]{eltanbouly-etal-2025-trates}
Sohaila Eltanbouly, Salam Albatarni, and Tamer Elsayed. 2025.
\newblock \href {https://doi.org/10.18653/v1/2025.findings-acl.1054} {{TRATES}: Trait-specific rubric-assisted cross-prompt essay scoring}.
\newblock In \emph{Findings of the Association for Computational Linguistics: ACL 2025}, pages 20528--20543, Vienna, Austria. Association for Computational Linguistics.

\bibitem[{Ghazawi and Simpson(2024)}]{ghazawi2024automated}
Rayed Ghazawi and Edwin Simpson. 2024.
\newblock Automated essay scoring in {A}rabic: a dataset and analysis of a {BERT}-based system.
\newblock \emph{arXiv preprint arXiv:2407.11212}.

\bibitem[{Ghazawi and Simpson(2025)}]{ghazawi2025well}
Rayed Ghazawi and Edwin Simpson. 2025.
\newblock How well can {LLMs} grade essays in {A}rabic?
\newblock \emph{Computers and Education: Artificial Intelligence}, page 100449.

\bibitem[{Gong et~al.(2021)Gong, Hu, Song, Fu, Sheng, Zhu, Wang, and Liu}]{gong2021iflyea}
Jiefu Gong, Xiao Hu, Wei Song, Ruiji Fu, Zhichao Sheng, Bo~Zhu, Shijin Wang, and Ting Liu. 2021.
\newblock \href {https://aclanthology.org/2021.acl-demo.29/} {{IF}ly{EA}: A {C}hinese essay assessment system with automated rating, review generation, and recommendation}.
\newblock In \emph{Proceedings of the 59th Annual Meeting of the Association for Computational Linguistics and the 11th International Joint Conference on Natural Language Processing: System Demonstrations}, pages 240--248, Online. Association for Computational Linguistics.

\bibitem[{Hirao et~al.(2020)Hirao, Arai, Shimanaka, Katsumata, and Komachi}]{hirao2020automated}
Reo Hirao, Mio Arai, Hiroki Shimanaka, Satoru Katsumata, and Mamoru Komachi. 2020.
\newblock \href {https://aclanthology.org/2020.lrec-1.157/} {Automated essay scoring system for nonnative {J}apanese learners}.
\newblock In \emph{Proceedings of the Twelfth Language Resources and Evaluation Conference (LREC 2020)}, pages 1250--1257, Marseille, France. European Language Resources Association (ELRA).

\bibitem[{Kumar et~al.(2022)Kumar, Mathias, Saha, and Bhattacharyya}]{manyhands}
Rahul Kumar, Sandeep Mathias, Sriparna Saha, and Pushpak Bhattacharyya. 2022.
\newblock \href {https://doi.org/10.18653/v1/2022.naacl-main.106} {Many hands make light work: Using essay traits to automatically score essays}.
\newblock In \emph{Proceedings of the 2022 Conference of the North American Chapter of the Association for Computational Linguistics: Human Language Technologies}, pages 1485--1495, Seattle, United States. Association for Computational Linguistics.

\bibitem[{Li and Ng(2024)}]{li-ng-2024-conundrums}
Shengjie Li and Vincent Ng. 2024.
\newblock \href {https://doi.org/10.18653/v1/2024.acl-long.414} {Conundrums in cross-prompt automated essay scoring: Making sense of the state of the art}.
\newblock In \emph{Proceedings of the 62nd Annual Meeting of the Association for Computational Linguistics (Volume 1: Long Papers)}, pages 7661--7681, Bangkok, Thailand. Association for Computational Linguistics.

\bibitem[{Liu et~al.(2025)Liu, Litman, Wang, Li, Gobat, Matsumura, and Correnti}]{liu-etal-2025-erevise}
Zhexiong Liu, Diane Litman, Elaine~L Wang, Tianwen Li, Mason Gobat, Lindsay~Clare Matsumura, and Richard Correnti. 2025.
\newblock \href {https://doi.org/10.18653/v1/2025.naacl-demo.18} {e{R}evise+{RF}: A writing evaluation system for assessing student essay revisions and providing formative feedback}.
\newblock In \emph{Proceedings of the 2025 Conference of the Nations of the Americas Chapter of the Association for Computational Linguistics: Human Language Technologies (System Demonstrations)}, pages 173--190, Albuquerque, New Mexico. Association for Computational Linguistics.

\bibitem[{Machhout and Zribi(2024)}]{machhout2024enhanced}
Rim~Aroua Machhout and Chiraz Ben~Othmane Zribi. 2024.
\newblock \href {https://doi.org/10.5171/2024.176992} {Enhanced {BERT} approach to score {A}rabic essay’s relevance to the prompt}.
\newblock \emph{Communications of the IBIMA}, 2024.

\bibitem[{Mahmoud et~al.(2024)Mahmoud, Nabil, and Torki}]{mahmoud2024automatic}
Somaia Mahmoud, Emad Nabil, and Marwan Torki. 2024.
\newblock Automatic scoring of {A}rabic essays: A parameter-efficient approach for grammatical assessment.
\newblock \emph{IEEE Access}.

\bibitem[{Ormerod(2022)}]{ormerod2022mapping}
Christopher~Michael Ormerod. 2022.
\newblock Mapping between hidden states and features to validate automated essay scoring using deberta models.
\newblock \emph{Psychological Test and Assessment Modeling}, 64(4):495--526.

\bibitem[{Ridley et~al.(2021)Ridley, He, Dai, Huang, and Chen}]{ridley2021automated}
Robert Ridley, Liang He, Xin-yu Dai, Shujian Huang, and Jiajun Chen. 2021.
\newblock Automated cross-prompt scoring of essay traits.
\newblock In \emph{Proceedings of the AAAI conference on artificial intelligence}, volume~35, pages 13745--13753.

\bibitem[{Sayed et~al.(2025)Sayed, Eltanbouly, Bashendy, and Elsayed}]{sayed2025feature}
Marwan Sayed, Sohaila Eltanbouly, May Bashendy, and Tamer Elsayed. 2025.
\newblock \href {https://doi.org/10.18653/v1/2025.arabicnlp-main.19} {Feature engineering is not dead: A step towards state of the art for {A}rabic automated essay scoring}.
\newblock In \emph{Proceedings of the Arabic Natural Language Processing Conference (ArabicNLP 2025)}, pages 231--245, China.

\bibitem[{Taghipour and Ng(2016)}]{taghipour-ng-2016-neural}
Kaveh Taghipour and Hwee~Tou Ng. 2016.
\newblock \href {https://doi.org/10.18653/v1/D16-1193} {A neural approach to automated essay scoring}.
\newblock In \emph{Proceedings of the 2016 Conference on Empirical Methods in Natural Language Processing}, pages 1882--1891, Austin, Texas. Association for Computational Linguistics.

\bibitem[{Team et~al.(2025)Team, Abbas, Ahmad, Alam, Altinisik, Asgari, Boshmaf, Boughorbel, Chawla, Chowdhury et~al.}]{team2025fanar}
Fanar Team, Ummar Abbas, Mohammad~Shahmeer Ahmad, Firoj Alam, Enes Altinisik, Ehsannedin Asgari, Yazan Boshmaf, Sabri Boughorbel, Sanjay Chawla, Shammur Chowdhury, and 1 others. 2025.
\newblock Fanar: An {A}rabic-centric multimodal generative ai platform.
\newblock \emph{arXiv preprint arXiv:2501.13944}.

\bibitem[{Xie et~al.(2022)Xie, Cai, Kong, Zhou, and Qu}]{npcr}
Jiayi Xie, Kaiwei Cai, Li~Kong, Junsheng Zhou, and Weiguang Qu. 2022.
\newblock \href {https://aclanthology.org/2022.coling-1.240} {Automated essay scoring via pairwise contrastive regression}.
\newblock In \emph{Proceedings of the 29th International Conference on Computational Linguistics}, pages 2724--2733, Gyeongju, Republic of Korea. International Committee on Computational Linguistics.

\bibitem[{Zhang et~al.(2025)Zhang, Deng, Dong, Zhao, Liu, and Cui}]{pairwise-holistic-2025}
Chunyun Zhang, Jiqin Deng, Xiaolin Dong, Hongyan Zhao, Kailin Liu, and Chaoran Cui. 2025.
\newblock \href {https://doi.org/10.1016/j.eswa.2024.125924} {Pairwise dual-level alignment for cross-prompt automated essay scoring}.
\newblock \emph{Expert Systems with Applications}, 265:125924.

\end{thebibliography}

% \newpage
\appendix

% \section{Appendix}

\section{Writing Prompt and Rubrics Examples}
\label{rubric-examples}
% [To be organized better]
% Example of a trait within a rubric: 
% \begin{description}\label{rubricExample}
%     \item[Trait] Vocabulary
%     \item[Range] 0-5
%     \item[Description] 5 - Excellent grammar, only two mistakes or fewer present. 4 -  Very good grammar, with only three to five mistakes present. 3 - Good grammar...
% \end{description}

Figure \ref{fig:prompt-example} and Table \ref{tab:rubric-example} illustrate the English translation of an example writing prompt and the corresponding rubric for the vocabulary trait drawn from LAILA dataset.
% . These examples are drawn from LAILA dataset, which was used to train our scoring models.
% These examples are presented to clarify the assignment design and the rubric-based assessment framework adopted in our system.

\begin{figure}[ht]
    \centering
    \begin{minipage}{\linewidth}
        \begin{tcolorbox}[colback=gray!5, colframe=black, boxrule=0.5pt, left=2pt, right=2pt, top=2pt, bottom=2pt]
{\fontsize{8}{10}\selectfont

% \begin{arabtex} 
% على الرغم من أهمية وسائل التواصل الاجتماعي وتأثيرها الإيجابي في توسيع دائرة العَلاقات الاجتماعية، إلا أنَّ البعض يرى فيها تهديدًا للعَلاقات الأسرية؛ وإصابة أفرادِ الأسرة بالعزلة والانطواء، كما أنها قد تتسبب في تدني التحصيل الدراسي لدى بعض الطلاب. كيف ترى هذه القضية؟
% اكتب مقالًا مكوَّنًا من ثلاثمائة (300) كلمة لتقنع به القارئ بوجهة نظرك موظّفًا الأدلة والحجج الدّاعمة لهذا الرأي، ومراعيًا أساليب الإقناع، ومستخدمًا علامات الترقيم وأدوات الربط المناسبة.
% \end{arabtex}

\color{darkgray}
{\fontsize{10}{10}\selectfont
% Despite the importance of social media and its positive impact on expanding the circle of social relationships, some see it as a threat to family relationships, causing family members to become isolated and introverted. Additionally, it may contribute to lower academic achievement among some students. How do you see this issue?
Do you agree or disagree with the following statement? Telephones and emails have made face-to-face communication between people much less likely.
Write a 500-word essay to persuade the reader of your point of view, employing supporting evidence and arguments, considering persuasive techniques, and using appropriate punctuation and linking words.

}

}
\end{tcolorbox}
        \captionof{figure}{Example of a writing prompt from LAILA.}
\label{fig:prompt-example}
    \end{minipage}
\end{figure}

\begin{table}[h]
\small
\centering
\begin{tabularx}{\linewidth}{|p{0.6cm}|X|}
\hline
\textbf{Score} & \textbf{Description} \\ \hline
1 & Use of a limited range of vocabulary and phrases that do not make sense together, with repetition and lexical errors, and generally inappropriate vocabulary that obscures meaning. \\ \hline
2 & Use of a basic range of vocabulary, with repetition, lexical errors, and many inappropriate choices that may obscure meaning. \\ \hline
3 & Use of a sufficient range of vocabulary, with some repetition and lexical errors, and a small number of inappropriate choices that may obscure meaning. \\ \hline
4 & Use of a good and appropriate range of vocabulary with few lexical errors, occasional inappropriate choices that do not affect meaning, and occasional use of idiomatic expressions. \\ \hline
5 & Use of a broad, correct, and appropriate range of vocabulary with few errors, demonstrating good knowledge of idiomatic expressions and awareness of implicit levels of meaning. \\ \hline
% \multicolumn{2}{|p{7.25cm}|}{\textbf{Note:} A score of 0 is assigned if the student does not attempt the task, provides a response below the performance level described for score 1, or submits content unrelated to the prompt.} \\ 
% \hline
\end{tabularx}
\caption{Rubric for the \textbf{vocabulary} trait.}
\label{tab:rubric-example}
\end{table}

\begin{table}[h!]
\small
\centering
\begin{tabular}{llc}
\hline
\textbf{Model} & \textbf{Hyperparameter} & \textbf{Value} \\
\hline
\textbf{RF} & max depth & 6 \\
 & max features & 0.1 \\
 & max samples & 0.1 \\
 & min samples split & 5 \\
 & \#estimators & 200 \\
 & feature selection threshold & 0.4 \\
 \hline
\textbf{XGB} & \#estimators & 300 \\
 & max depth & 3 \\
 & learning rate & 0.1 \\
 & subsample & 0.9 \\
 & feature selection threshold & 0.1 \\
 \hline
\textbf{NN} & \#epochs & 50 \\
 % & early stopping patience & 10 \\
 & \#layers & 2 \\
 & dropout & 0.3 \\
 & loss & MSE \\
 & optimizer & AdamW \\
 & learning rate & 1e-4 \\
 & batch size & 16 \\
 & hidden layer widths & 256 \\
 & feature selection threshold & 0.5 \\
 \hline
\end{tabular}
\caption{The hyperparameter values used to train the deployed RF, XGB, and NN models}
\label{tab:FB-hyperparameters}
\end{table}

\begin{table*}[]
\small
\centering
\begin{tabular}{lcccccccc}
\hline
 & \textbf{REL} & \textbf{ORG} & \textbf{VOC} & \textbf{STY} & \textbf{DEV} & \textbf{MEC} & \textbf{GRM} & \textbf{HOL} \\
 \hline
loss & WMSE & WMSE & WMSE & WMSE & WMSE & WMSE & WMSE & MSE \\
learning rate & 0.001 & 0.0001 & 0.0001 & 0.01 & 0.001 & 0.01 & 0.0001 & 0.01 \\
hidden layer number \& widths & 64 & 64 & 32 & 64 & 32, 16 & 64 & 64 & 32, 32 \\
weight decay & 1e-6 & 1e-4 & 1e-4 & 1e-4 & 1e-4 & 1e-6 & 1e-4 & 1e-5 \\
dropout & 0.1 & 0 & 0 & 0 & 0 & 0.2 & 0.2 & 0.2 \\
feature selection threshold & 0.5 & 0.5 & 0.4 & 0.3 & 0.5 & 0.5 & 0.5 & 0.5 \\
\#epochs & 80 & 87 & 85 & 36 & 74 & 52 & 76 & 69 \\
\hline
\end{tabular}
\caption{The hyperparameter values used to train TRATES. WMSE refers to the weighted MSE.}
\label{tab:TRATES-hyperparameters}
\end{table*}

\section{Scoring Interface}
\label{appendix:scoring-interface}

\begin{figure*}[ht]
    \centering
    \includegraphics[width=\textwidth]{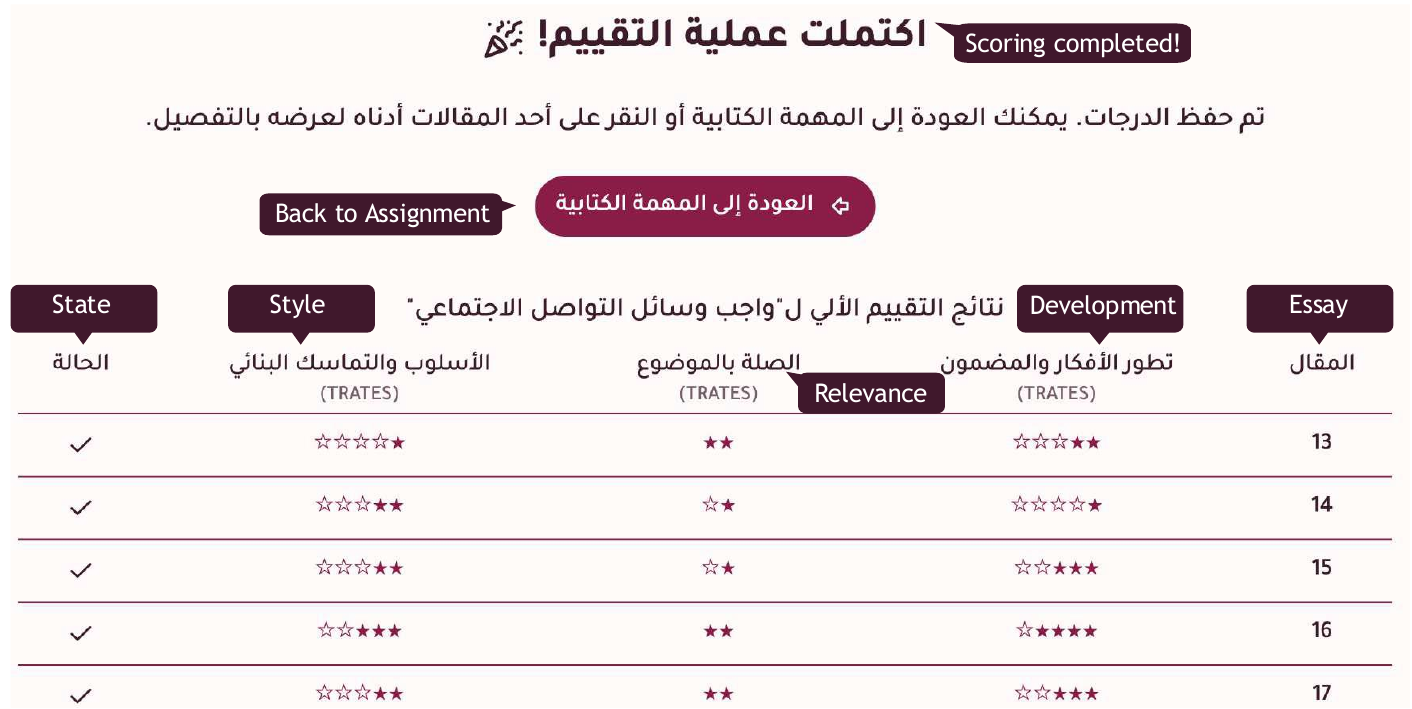}
    \caption{Real-time scoring view of 5 essays evaluated across three traits: development, relevance, and style.}
    \label{fig:scoring-view}
\end{figure*}

Figure~\ref{fig:scoring-view} shows the real-time scoring view after a batch of essays has been scored for three traits: development, relevance, and style. Each row corresponds to an essay, with columns showing the scores assigned by the selected model for each trait. Scores are rendered as star ratings reflecting the score range of the trait: 0-2 for relevance and 0-5 for development and style. The State column indicates whether scoring is completed successfully, and the model used for each trait is shown beneath the trait name. From this view, the user can open an individual essay for inspection and possible manual edits of the scores, or return to the assignment to launch additional scoring batches.

\section{Implementation Details of Scoring Models} \label{appendix:immplementation-detail}

All models are trained on LAILA dataset using the released cross-prompt splits,\footnote{\url{https://gitlab.com/bigirqu/laila}}
following a leave-one-prompt-out cross-validation setup. Hyperparameter tuning is performed, and the best configuration is selected based on development set performance, after which results are reported on the unseen target prompt.
For training the deployment models, the models are retrained on the full dataset using the most frequently selected hyperparameter configurations across the individual folds.

\subsection{Feature-based Models} \label{appendix:feature-based}

The three feature-based models are trained in a multi-task setup, where all traits are predicted simultaneously. For model training, we adopt the 816 handcrafted features introduced by \citet{sayed2025feature} for Arabic AES, covering surface, readability, lexical, semantic, and syntactic features. 
To reduce noise, we performed feature selection based on Pearson and Spearman correlations \cite{li-ng-2024-conundrums}, retaining features whose absolute correlation with any trait exceeded a predefined threshold. For the implementation of RF and XBG, we used the sklearn library\footnote{\url{https://scikit-learn.org/}} and the XGBoost library,\footnote{\url{https://xgboost.readthedocs.io/en/stable/}} while the NN is implemented using Pytorch. Table~\ref{tab:FB-hyperparameters} summarizes the hyperparameters used to train the deployment models.

For deployment, 22 features that require pre-trained transformer models for extraction are removed. These features were found to contribute minimally to overall performance while increasing scoring time; therefore, they are excluded to improve efficiency.

\subsection{TRATES Model} \label{appendix:TRATES}
TRATES pipeline consists of three stages \cite{eltanbouly-etal-2025-trates}. In the feature generation stage, we use \texttt{Fanar-1-9B-Instruct}\footnote{\url{https://huggingface.co/QCRI/Fanar-1-9B-Instruct}}
to generate trait-specific features from the rubrics. For holistic scoring, LAILA dataset does not provide a separate rubric for the holistic score. Therefore, to generate features for holistic scoring, Fanar is prompted with an aggregated list of all trait features and asked to generate a set of 10 features covering the main scoring aspects.
In the feature extraction stage, Fanar is also used, with extraction optimized by querying all rubric-based features for a given essay in a single call rather than a separate call for each feature.
In the final stage, which combines trait-specific features with engineered linguistic features, we use the same feature set adopted for the feature-based models, along with the same feature selection method. This stage trains trait-specific neural network regression models that take the combined feature set as input and predict the corresponding trait score. 
The hyperparameters used for the deployed models are presented in Table~\ref{tab:TRATES-hyperparameters}.

% We reimplemented two SOTA English AES models for Arabic. The first model is TRATES~\cite{eltanbouly-etal-2025-trates}, which generates trait-specific features from essays using an LLM, which are then combined with engineered linguistic features for scoring using a simple neural network. In our adaptation, both feature generation and extraction use the open-source \texttt{Fanar-1-9B},\footnote{\url{https://huggingface.co/QCRI/Fanar-1-9B}} with extraction optimized by querying all rubric-based questions for an essay in a single call. 
% % Engineered features include length, readability, POS, complexity, sentiment, and prompt-specific features derived from the Arabic feature set.
% The second model is MOOSE~\cite{chen-etal-2025-mixture-ordered}, which is a cross-prompt model combining three experts for trait prediction, pairwise essay ranking, and essay–prompt relevance via a mixture-of-experts. For Arabic, we adapted MOOSE using AraBERT \cite{antoun2020arabert} and the same set of Arabic handcrafted features \cite{sayed2025feature}.
% Both adaptations preserve the original methodology while ensuring suitability for deployment on \sys{} for Arabic AES.

\begin{table}[]
\small
\centering
\begin{tabular}{lc}
\hline
\textbf{Hyperparameter} & \textbf{Value} \\
\hline
batch\_size & 4 \\
\#epoch & 15 \\
encoder & AraBERT \\
learning\_rate & 2e-5 \\
\hline
\end{tabular}
\caption{The hyperparameter values used to train the deployed MOOSE model.}
\label{tab:moose-hyperparameters}
\end{table}

\subsection{MOOSE Model} \label{appendix:MOOSE}
MOOSE \cite{chen-etal-2025-mixture-ordered} is a multi-trait, cross-prompt essay scoring model composed of three specialized experts: a scoring expert, which learns inherent scoring cues from essays; a ranking expert, which evaluates the relative quality across different essays; and an adherence expert, which estimates the degree of prompt adherence.
To adapt MOOSE for Arabic, we replaced BERT with AraBERT and incorporated the engineered Arabic feature set for both essay and prompt representations \cite{sayed2025feature}. For implementation, we used the official implementation released by the authors.\footnote{\url{https://github.com/antslabtw/MOOSE-AES}}
During hyperparameter tuning, the learning rate was explored over the values 1e-4 and 2e-5, while all other architectural parameters were kept consistent with those reported in the original study. The final hyperparameters are reported in Table~\ref{tab:moose-hyperparameters}.

\end{document}